\documentclass[10pt,twocolumn,letterpaper]{article}

\usepackage{cvpr}              

\usepackage{pifont}
\usepackage{makecell}
\usepackage{relsize}
\usepackage{appendix}
\usepackage{float}
\usepackage{ragged2e}
\usepackage{tikz}
\usepackage{subcaption}

\newcommand{\circlednumber}[1]{%
  \tikz[baseline=(char.base)]{
    \node[shape=circle, draw, inner sep=0.5pt, minimum size=0.4em] (char) {#1};}}

\newcommand{\modelname}{SGC-Net }
\usepackage{xspace}

\usepackage[accsupp]{axessibility}

\usepackage{lineno}

\input{def.set}
\definecolor{cvprblue}{rgb}{0.21,0.49,0.74}
\usepackage[pagebackref,breaklinks,colorlinks,allcolors=cvprblue]{hyperref}

\def\paperID{5342} 
\def\confName{CVPR}
\def\confYear{2025}

\title{SGC-Net: Stratified Granular Comparison Network for Open-Vocabulary HOI Detection}

\author{Xin Lin \quad Chong Shi \quad Zuopeng Yang\thanks{Corresponding author.}  \quad Haojin Tang\textsuperscript{\thefootnote}  \quad Zhili Zhou\\
 Guangzhou University \quad\\
{{\tt\small linxin94@gzhu.edu.cn,\quad shichong@e.gzhu.edu.cn,\quad
yzpeng@gzhu.edu.cn,\quad}} \\
{{\tt\small tanghaojin@gzhu.edu.cn,\quad
\texttt{zhou\_zhili}@163.com  }}}

\begin{document}
\maketitle
\begin{abstract}

Recent open-vocabulary human-object interaction (OV-HOI) detection methods primarily rely on large language model (LLM) for generating auxiliary descriptions and leverage knowledge distilled from CLIP to detect unseen interaction categories. Despite their effectiveness, these methods face two challenges: (1) feature granularity deficiency, due to reliance on last layer visual features for text alignment, leading to the neglect of crucial object-level details from intermediate layers; (2) semantic similarity confusion, resulting from CLIP's inherent biases toward certain classes, while LLM-generated descriptions based solely on labels fail to adequately capture inter-class similarities. To address these challenges, we propose a stratified granular comparison network. First, we introduce a granularity sensing alignment module that aggregates global semantic features with local details, refining interaction representations and ensuring robust alignment between intermediate visual features and text embeddings. Second, we develop a hierarchical group comparison module that recursively compares and groups classes using LLMs, generating fine-grained and discriminative descriptions for each interaction category. Experimental results on two widely-used benchmark datasets, SWIG-HOI and HICO-DET, demonstrate that our method achieves state-of-the-art results in OV-HOI detection. Codes will be released on \href{https://github.com/Phil0212/SGC-Net}{GitHub}.
\end{abstract}    
\section{Introduction}
\label{sec:intro}

Human-Object Interaction (HOI) detection aims to localize human-object pairs and recognize their interactions, providing an efficient way for human-centric scene understanding. It plays a crucial role in various computer vision tasks, such as assistive robotics \cite{mascaro2023hoi4abot,massardi2020parc} and video analysis \cite{xi2023open,li2024ripple}. 
Recently, the emerging field of open-vocabulary HOI (OV-HOI) detection, which broadens the scope of HOI detection by recognizing and associating objects beyond predefined categories, has gained increasing attention.

\begin{figure}[t]
\begin{center}
    \centering
    \includegraphics[scale=0.65]{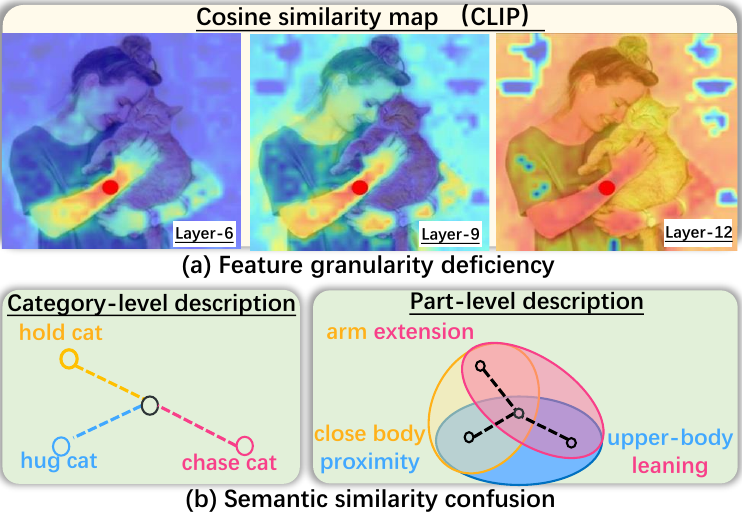}
    \captionof 
    {figure}{(a) The last layer capture high-level global semantics but contain fewer low-level local details compared to intermediate layers. The red dot marks the selected patch. (b) Category-level and part-level descriptions overlook inter-class similarity, leading to difficulty in distinguishing semantically similar classes.
    }\label{intro1}
\end{center}
\end{figure}

OV-HOI detection has made notable progress, largely due to the success of vision-language models (VLMs) like CLIP~\cite{radford2021learning}. Unlike zero-shot learning, which prohibits the access of any unannotated unseen classes during training, open-vocabulary learning leverages class name embeddings from VLMs for auxiliary supervision~\cite{wu2024towards, zhu2024survey}. Therefore, existing CLIP-based HOI detection methods naturally fall under OV-HOI detection. However, these methods face challenges in open-world scenarios because they rely on pretrained object detectors, and using category names as classifiers struggles to capture the variability of HOIs. To overcome these challenges, recent works~\cite{wang2022learning, lei2024exploring} have removed the need for pretrained detectors by extracting HOI features directly through image encoders while incorporating language priors from LLMs via text encoders.

Despite these advancements, current OV-HOI detection methods without pretrained object detectors still suffer from feature granularity deficiency. This issue arises because CLIP, trained with image-level alignment, produces globally aligned image-text features lacking the local detail needed for effective OV-HOI detection. As shown in Figure~\ref{intro1}(a), the last layer focuses on high-level semantic features but neglects low-level local details like arm and face postures, compared to intermediate layers. Additionally, feature dissimilarity between shallow and deep layers increases with network depth, posing significant challenges in aligning intermediate visual features with text embeddings. Moreover, OV-HOI methods that rely on category-level or part-level text classifiers are prone to semantic confusion due to several inherent limitations. First, CLIP's training on large-scale, long-tail datasets tends to introduce biases towards certain classes \cite{wang2022debiased}. For example, as shown on the left side of Figure~\ref{intro1}(b), CLIP often confuses ``hug cat" with ``hold cat." Second, descriptions generated by LLMs based solely on labels may fail to distinguish between semantically similar classes. As illustrated on the right side of Figure~\ref{intro1}(b), LLMs produce similar part-level descriptions like ``arm extension" for both ``hold cat" and ``chase cat".

To address the aforementioned challenges, we propose an end-to-end OV-HOI detection network named SGC-Net. First, we design the granularity sensing alignment (GSA) module, leveraging multi-granularity visual features from CLIP to improve the transferability for novel classes. The primary challenge in this process is effectively aggregating global semantic features with local details to refine coarse interaction representations while ensuring alignment between intermediate visual features and text embeddings. Accordingly, we introduce a block partitioning strategy that groups layers based on their relative distances. Specifically, we divide the visual encoder of CLIP into several blocks, ensuring a little variation in the features within each block. To align local details more precisely, we perform a fusion of multi-granularity features within each block, assigning distinct and trainable Gaussian weights to each layer. Once these intra-block features are fused, we proceed to aggregate the fused features across all blocks to maintain global semantic consistency. Furthermore, to preserve the pre-trained alignment between text and visual features in CLIP, we employ visual prompt tuning \cite{li2024cascade,zhou2023zegclip,ding2022decoupling} by adding learnable tokens to the visual features of each layer.

Secondly, we propose the hierarchical group comparison (HGC) module to effectively differentiate semantically similar categories by recursively grouping and comparing classes via LLMs. Specifically, we utilize clustering algorithms (\eg, $K$-means \cite{macqueen1967some}) for grouping. Even with extensive prior knowledge of LLMs, comparing numerous categories for OV-HOI detection remains challenging due to the quadratic growth of the description matrix with the number of categories. To address this challenge, we adopt group-specific comparison strategies to improve efficiency. For smaller groups, we directly query LLMs for comparative descriptions, whereas for larger groups, we summarize the group's characteristics using LLMs and incorporate these summaries into new queries to highlight distinctive features of each class within the group. By recursively construct a class hierarchy, we can classify HOI by descending from the top to the bottom of the hierarchy, comparing HOI and text embeddings at each level.

In summary, the innovation of the proposed SGC-Net is three-fold: (1) The GSA module effectively aggregates multi-granular features while aligning the local details and global semantic features of OV-HOI; (2) The HGC module iteratively generates discriminative descriptions to refine the classification boundaries of labels; (3) The efficacy of the proposed SGC-Net is evaluated on two widely-used OV-HOI detection benchmark datasets, \ie., HICO-DET~\cite{chao2018learning} and SWIG-HOI~\cite{wang2021discovering}. Experimental results show that our SGC-Net consistently outperforms state-of-the-art methods.
\section{Related Work}
\label{sec:related_work}

\subsection{HOI Detection}

Existing HOI detection works can be divided into two categories, two-stage methods \cite{cao2023re, gao2018ican, gao2020drg, lei2023efficient, li2019transferable, park2023viplo,tamura2021qpic, zhang2022efficient, zhang2023exploring} and one-stage methods \cite{chen2021reformulating, gkioxari2018detecting, kim2020uniondet, kim2022mstr, liao2020ppdm, tu2023agglomerative, zhong2021glance}. Two-stage methods typically use an object detector to recognize humans and objects, followed by specialized modules to associate humans with objects and identify their interactions, \eg., multi-steams \cite{gkioxari2018detecting,gupta2019no,li2020pastanet}, graphs \cite{qi2018learning,xu2019learning,gao2020drg, zhang2021spatially} or compositional learning \cite{hou2020visual,hou2021detecting,hou2021affordance}. In contrast, the one-stage approach detects human-object pairs and their interactions simultaneously, without requiring stepwise processing. In particular, RLIP \cite{zhou2019relation} proposes a pre-training strategy for HOI detection based on image captions. PPDM \cite{liao2020ppdm} reformulates the HOI detection task as a point detection and matching problem and achieves simultaneous object and interaction detection. However, existing HOI detection approaches are constrained by a closed-set assumption, which restricts their ability to recognize only predefined interaction categories. In contrast, our work aims to detect and recognize HOIs in the open-vocabulary scenario.

\subsection{CLIP-based Open Vocabulary HOI Detection}

To alleviate the closed-set limitation, many studies \cite{lei2024exploring, Wang_2022_CVPR,mao2023clip4hoi,ning2023hoiclip,liao2022gen, gao2024contextual, yang2024open} have been devoted to OV-HOI detection, aiming to identify both base and novel categories of HOIs while only base categories are available during training. Specifically, they transfer knowledge from the large-scale visual-linguistic pre-trained model CLIP to enhance interaction understanding. For instance, GEN-VLKTs \cite{liao2022gen} and HOICLIP \cite{ning2023hoiclip} convert HOI labels into phrase descriptions to initialize the classifier and extract visual features from the CLIP image encoder to guide the interaction visual feature learning. MP-HOI \cite{yang2024open} incorporates the visual prompts into language-guided-only HOI detectors to enhance its generalization capability. THID \cite{wang2022learning} proposes a HOI sequence parser to detect multiple interactions. CMD-SE~\cite{lei2024exploring} detects HOIs at different distances using distinct feature maps. However, the adopted loss function in ~\cite{lei2024exploring} involves minimizing differences between continuous and discrete variables therefore hard to optimize. Besides, aligning intermediate features with text embeddings may break the alignment in pre-trained CLIP due to the substantial differences. The proposed approach is a method free from pretrained object detectors but has the following advantages compared with existing works. First, it aggregates multi-granularity features from the visual encoder more appropriately. Second, it is easy to optimize and deploy to existing models.

\subsection{Leverage LLM for visual classification}
Recently, LLMs~\cite{brown2020language,biderman2023pythia,driess2023palm,iyer2022opt} have made remarkable progress, revolutionizing natural language processing tasks with hundreds of billions of parameters and techniques such as reinforcement learning. Existing works ~\cite{cao2024detecting,jin2024llms,unal2023weakly,zhou2022learning} have shown their effectiveness in generating comprehensive descriptions, particularly in classification and detection tasks. Specifically, Menon and Vondrick \cite{menon2022visual} generate textual descriptions directly from labels to assist VLMs in image classification. I2MVFormer \cite{naeem2023i2mvformer} leverages LLM to generate multi-view document supervision for zero-shot image classification. ContexDET \cite{zang2024contextual} utilizes contextual LLM tokens as conditional object queries to enhance the visual decoder for object detection. RECODE \cite{li2024zero} leverages LLMs to generate descriptions for different components of relation categories. However, descriptions generated solely by LLMs from labels often tend to be generic and lack sufficient discriminability among semantically similar classes. To address this issue, our work leverages LLM to generate text descriptions by comparing different classes to reduce inter-class similarity and make the decision boundary of each class more compact.

\section{Method}

\begin{figure*}[t]
  \centering
   \includegraphics[width=1.0\linewidth]{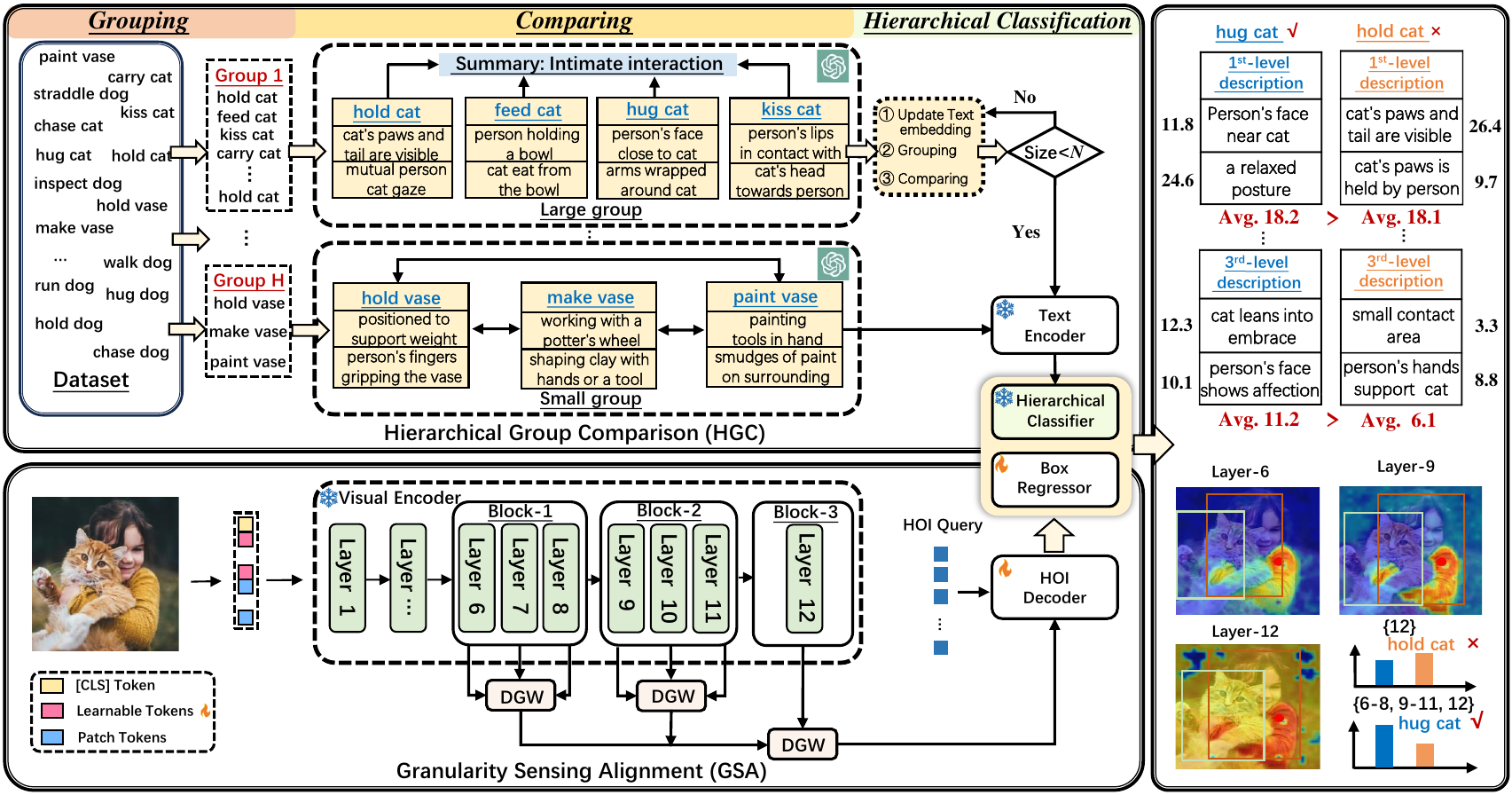}
   \vspace{-5mm}
   \caption{The framework of SGC-Net. It eliminates the need for a pretrained object detector and includes two new modules for OV-HOI detection: (1) The GSA module partitions the CLIP visual encoder into blocks, aggregates features via DGW, and integrates an HOI decoder for fine-grained HOI representations. (2) The HGC module uses LLM to recursively construct class hierarchy, enabling HOI classification by traversing the hierarchy from top to bottom and comparing HOI representations with text embeddings at each level. }
   \label{fig:pipeline}
\end{figure*}

This section provides details of the proposed stratified granular comparison network (SGC-Net). Specifically, we first describe the network architecture, followed by an explanation of the training and inference pipeline. As shown in Figure~\ref{fig:pipeline}, SGC-Net comprises a granularity sensing alignment (GSA) module and a hierarchical group comparison (HGC) module. The GSA module enhances coarse HOI representations by utilizing distance-Aware Gaussian weighting (DGW) and visual prompt tuning to combine multi-granularity features from CLIP's image encoder. The HGC module refines decision boundaries for semantically similar classes by recursively comparing and grouping them using LLM-derived knowledge. In the below, we will describe these two components sequentially.

\subsection{Granularity Sensing Alignment}

Existing OV-HOI methods~\cite{ning2023hoiclip, liao2022gen, lei2024exploring} typically use features from the last layer of CLIP's image encoder to model HOIs. While these deep features effectively capture high-level semantics, they often lack the local details necessary for HOI detection. One approach to address this issue is to aggregate feature maps from different levels and feed them into an interaction decoder. However, the significant differences between shallow and deep features can disrupt the pre-trained vision-language alignment, ultimately reducing CLIP's zero-shot capability for OV-HOI tasks.

To address this issue, we propose the GSA module, which effectively captures fine-grained details in human-object interactions while preserving the vision-language alignment in the pre-trained CLIP model. Specifically, we first divide the CLIP’s image encoder into $S$ blocks, ensuring a little variation in the features within each block. Then, within each block, distinct distance-aware Gaussian weights are assigned to the transformer layers based on their relative neighborhood distances. These weights are trainable, enabling the model to adaptively learn layer- and block-specific information from the training data. Finally, we aggregate the features across different blocks. For each block containing 
$d$ transformer layers, the aggregated feature $\bZ$ can be represented as follows:

\begin{equation}\label{fusion}
\begin{aligned}
\alpha_{l}^s &= \exp\left(-\frac{1}{2} \frac{(d - l)^2}{\sigma^2}\right), \quad l \in [1, d],\\
\bZ &= \sum_{s=1}^{S} \alpha_s \left(\sum_{l=1}^{d}\alpha_{l}^s {\bF}_{\!l}\right)
\end{aligned}
\end{equation}
Here, $l$ represents the layer index. ${\bF}_{\!l}$ is a feature map from the $l$-th layer of CLIP image encoder. $\alpha_{l}^s$ and $\alpha_s$ denote the distance-aware Gaussian weights between layers within the same block and across different blocks, respectively. By appropriately setting variance parameter $\sigma$, the Eq.~\eqref{fusion} can assign a higher weight to nearby layers and a lower weight to distant ones, facilitating more effective and flexible integration of features across different depth levels. Notably, the last layer of CLIP is treated as a separate block with a large weight. This design aims to keep the original visual-language correlations established in the pre-trained model. Different blocks can capture varying levels of granularity. Specifically, the early blocks focus more on fine-grained local details, while the later block emphasizes coarse-grained global features.

To better align the intermediate visual features with the text embeddings, we employ visual prompt tuning~\cite{zhou2022learning} by introducing learnable tokens onto the visual features of each layer in the frozen encoder. During visual prompt tuning, the GSA module enables the gradients to be directly back-propagated to the middle layers of the visual encoder. This can promote the alignment of mid-layer features and text embeddings, substantially enhancing the similarity across different layers.

Compared with CMD-SE~\cite{lei2024exploring}, we effectively aggregate multi-granularity features from the CLIP image encoder using trainable Gaussian weights, which have the following advantages. First, it allows intermediate layer visual features to complement last layer features while preserving the visual-language associations of the pre-trained CLIP. This approach addresses the issue in CMD-SE~\cite{lei2024exploring}, where relying solely on discrete single-layer features for HOI modeling results in a loss of fine-grained information. Second, our method is easy to optimize. We eliminate the need to use loss functions that align discrete transformer layer indices with continuous human-object interaction distances, thereby avoiding the complex optimization processes.

Subsequently, following previous works \cite{lei2024exploring, liao2022gen, ning2023hoiclip}, the final interaction representation $\bX$ is formulated as follows: 
\begin{equation}\label{decoder}
\bX = \rm Dec \left( \bQ, \bZ \right),
\end{equation}
where $\bQ$ refers to HOI queries, while $\bZ$ is treated as both the key and value input to the HOI Decoder. The output $\bX$ is later passed to the bounding box regressor and classifier to predict the bounding boxes of human-object pairs and their interaction categories.

\subsection{Hierarchical Group Comparison}

After obtaining the fine-grained interaction representation by the GSA module, designing a discriminative classifier remains a challenging task. Existing methods \cite{ning2023hoiclip,liao2022gen,lei2024exploring} typically design classifiers with the manual prompt ``a photo of a person [verb] [object]". However, these classifiers primarily rely on category names, often neglecting the contextual information provided by language. Recent research \cite{lei2024exploring} has introduced the category descriptions produced by LLMs to enhance the model's generalization capability. However, these descriptions directly generated by LLMs from labels are often generic and semantically similar.

To address this issue, we propose the hierarchical group comparison (HGC) module. Inspired by the coarse-to-fine approach humans use to recognize objects, HGC employs three strategies: Grouping, Comparison, and Hierarchical Classification. It identifies semantically similar descriptions and refines decision boundaries from generic to more discriminative. Specifically, we first utilize LLMs to generate initial descriptions for each HOI category.

\vspace{1mm}
{\small\noindent{\texttt{Q: What features are useful to distinguish \underline{\{HOI category\}} in a photo?}}}

{\small\noindent{\texttt{A: -}}}
\vspace{1mm}

\noindent\textbf{Grouping}: Following~\cite{menon2022visual}, we employ the pretrained CLIP text encoder to map the LLM-generated descriptions into the latent space. Utilizing these features, we apply a clustering algorithm, such as $K$-means~\cite{macqueen1967some}, to perform grouping and identify semantic neighbors. Specifically, the value of $K$ is calculated as the number of categories divided by a predefined grouping threshold $N$. Even with extensive prior knowledge of LLMs, comparing a large number of categories presents a formidable challenge, as it needs to generate a comprehensive description matrix that grows quadratically with the number of categories. To address this issue, we employ tailored strategies for generating descriptions based on the group size, allowing for more efficient and effective comparisons. Specifically, we adopt summary-based comparison for larger groups and direct comparison for smaller ones.

\noindent\textbf{Comparing}: 
For groups involving a large number of categories (\ie., exceeding half of the grouping threshold), we leverage the LLM to summarize their overall characteristics. By incorporating these summarized characteristics into a new query, we can generate a comparative description that captures the distinctive features and relationships among the categories. This approach enhances our ability to effectively compare and contrast the elements within a group.

\vspace{1mm}
{\small\noindent{\texttt{Q: Summarize the following interactions with one sentence: \underline{\{category list\}}?}}}

{\small\noindent{\texttt{A: \underline{\{subset description\}}}}}

\vspace{2mm}
{\small\noindent{\texttt{Q: What features are useful to distinguish \underline{\{HOI category\}} from \underline{\{subset description\}}?}}}

{\small\noindent{\texttt{A: -}}}
\vspace{1mm}

When the number of elements within a group is relatively small, a direct query to the LLM enables us to obtain an exceptionally comparative description. This approach capitalizes on the LLM’s capabilities to provide a comprehensive and insightful analysis, enhancing the comparative understanding of the elements within the group.

\vspace{1mm}
{\small\noindent{\texttt{Q: What features are useful to distinguish \underline{\{target category\}} from \underline{\{other categories\}} in a photo?}}}

{\small\noindent{\texttt{A: -}}}
\vspace{1mm}

Utilizing these detailed and comparative descriptions, we can achieve more nuanced text embedding. Subsequently, we can continue to perform clustering, followed by additional comparisons to derive new descriptors.

\noindent\textbf{Hierarchical Classification}:
After constructing the class hierarchy, we can obtain text features at multiple hierarchical levels. More specific, the hierarchical text features can be represented as $\mathcal{T}=\left\{{\bD_{1},\bD_{2},\cdots,\bD_{T}}\right\}$, where $T$ is the number of test classes. $\bD_{i}\in\mathbb{R}^{M_{j}\times{C}}$ represents the hierarchical text embeddings for category $i$. $C$ is the embedding dimension. $M_{j}$ is the number of comparative descriptions for category $i$. Accordingly, the cosine similarity score between the HOI feature $\bx$ and the $j$-th description of the $i$-th HOI category, denoted as $\bD_{i}^{j}$, can be expressed as follows:
\begin{equation}\label{eq:cosine-similarity}
 p_{i}^j= \bD_{i}^j \cdot \bx^T.
\end{equation}
However, unreliable low-level descriptions can introduce errors and redundancies, degrading the quality of high-level descriptions and ultimately impairing the model's classification performance. To address this issue, we define an iterative evaluator $\bu_{i}^{k}$ to evaluate the acceptability of the current $k$-th level description for the $i$-th category as follows:

\begin{equation}\label{mask}
\bu_{i}^{k}={\mathlarger{\mathbb{I}}}\left(p_{i}^{k+1} > p_{i}^{k}+\tau\right),
\end{equation}
where $\tau$ denotes the tolerance parameter, ${\mathbb{I}}$ represents indicator function. Eq.~\eqref{mask} ensures that a new score is merged only if a subsequent discriminative description yields a higher score than the current one. Subsequently, we obtain the running average of the longest sequence of monotonically increasing $p$ values as follows:
\begin{equation}
r(\bx,i) = \frac{{p}_{i}^{1} + \sum_{j=2}^{M_i} p_{i}^j \prod_{k=1}^{j-1} \bu_{i}^{k}}{1 + \sum_{j=2}^{M_i} \prod_{k=1}^{j-1} \bu_{i}^{k}}.
\end{equation}

Finally, the similarity score $s(\bx,i)$ between HOI feature $\bx$ and the $i$-th interaction label embedding is computed using a weighted average function, as follows:
\begin{equation}\label{final_s}
s(\bx,i) = (1 - \lambda) (\bp^1_i+\bt \cdot \bx^T) + \lambda \left( r(\bx, i) \right).
\end{equation}
Here, we apply prompt tuning~\cite{wang2022learning} with learnable text tokens $\bt$ to learn the sentence format rather than using manually defined context words. Besides, our fusion method takes into account the score of the initial description $\bp^1_i$ as an offset, and we introduce a constant hyperparameter $\lambda \in $ [0, 1] to balance the two terms.

\subsection{Training and Inference}        
\label{subsec:training}
In this subsection, we elaborate on the processes of training and inference of our model.

\noindent \textbf{Training.}
During the training stage, we follow the query-based methods~\cite{wang2022learning,liao2022gen, lei2024exploring} to assign a bipartite matching prediction with each ground-truth using the Hungarian algorithm~\cite{kuhn1955hungarian}.
The matching cost $\mathcal{L}$ for the matching process and the targeting cost for the training back-propagation share the same strategy, which is formulated as follows:

\begin{equation}\label{loss}
\mathcal{L} =\lambda_{b}\sum_{i \in\{h, o\}} \mathcal{L}_{b}^{i}+\lambda_{{iou }}\sum_{i \in\{h, o\}} \mathcal{L}_{{iou }}^{i}+\lambda_{cls} \mathcal{L}_{cls},
\end{equation}
where $\mathcal{L}_{b}$, $\mathcal{L}_{iou}$, and $\mathcal{L}_{cls}$ denote the box regression, intersection over union, and classification losses, respectively. During the training stage, we follow the query-based methods \cite{lei2024exploring, liao2022gen,ning2023hoiclip} to assign a bipartite matching prediction with each ground-truth using the Hungarian algorithm \cite{kuhn1955hungarian}. $\lambda_{b}$, $\lambda_{{iou }}$, and $\lambda_{cls}$ are the hyper-parameter weight.

\noindent \textbf{Inference.}
For each HOI prediction, including the bounding-box pair $(\hat{b_{h}^{i}}, \hat{b_{o}^{i}})$, the bounding box score $\hat{c_{i}}$ from the box regressor, and the interaction score $\hat{s_{i}}$ from the interaction classifier, the final score $\hat{s_{i}}'$ is computed as:
\begin{equation}
    \hat{s_{i}}' = \hat{s_{i}} \cdot \hat{c_{i}}^{\gamma}
\end{equation}
where $\gamma$ $\textgreater$ 1 is a constant used during inference to suppress
overconfident objects~\cite{zhang2021spatially,zhang2022efficient}.

\section{Experiment}
\label{sec:experiment}

\subsection{Experimental Setting}
\label{subsec:exp_setting}

\noindent \textbf{Datasets.}
Our experiments are conducted on two datasets: namely, SWIG-HOI~\cite{wang2021discovering} and HICO-DET~\cite{chao2018learning}. The SWIG-HOI dataset provides diverse human interactions with large-vocabulary objects, comprising 400 human actions and 1,000 object categories. During the testing, we utilize approximately 5,500 interactions, including around 1,800 interactions that are not found in the training set. The annotations of HICO-DET include 600 combinations of 117 human actions and 80 objects. Among the 600 interactions, follow \cite{hou2020visual,wang2022learning,lei2024exploring}, we simulate the open-vocabulary detection setting by holding out 120 rare interactions.

\noindent \textbf{Evaluation Metric.}
Following in ~\cite{chao2018learning,liu2022interactiveness,liao2022gen,wang2022learning,lei2024exploring}, we use the mean Average Precision (mAP) for evaluation. An HOI triplet prediction is classified as a true-positive example when the following criteria are satisfied: 1) The intersection over union of the human bounding box and object bounding box are larger than 0.5 \textit{w.r.t.} the GT bounding boxes; 2) the predicted interaction category is accurate.

\noindent \textbf{Implementation Details.}
We follow the settings of previous work \cite{wang2022learning,lei2024exploring} to build our model upon the pretrained CLIP. Specifically, for the visual encoder, we employ the ViT-B/16 version as our visual encoder and apply 12 learnable tokens in each layer to detect human-object interactions. For the text encoder, we introduce 8 prefix tokens and 4 conjunctive tokens to connect the words of human actions and objects for adaptively learning categories' information. Our model is optimized using AdamW with an initial learning rate of 10$^{-4}$, using 64 as the batch size for SWIG-HOI dataset and 32 for HICO-DET. We set the cost weights $\lambda_{b}$, $\lambda_{cls}$ and $\lambda_{{iou }}$ to 5, 2, and 5, respectively. The LLM we utilize is GPT-3.5. We set the hyperparameter $\lambda$ in Eq.~\eqref{final_s} as 0.5. In all experiments, the variance parameter $\sigma$ is set to 1, and the tolerance parameter $\tau$ is set to 0.

\begin{table}[t]
  \centering
  \begin{tabular}{@{}lcccc@{}}
    \toprule[2pt]
    \toprule
    Method & {\makecell[c]{Pretrained \\ Detector}} & Unseen & Seen & Full \\
     \hline
    FCL~\cite{hou2021detecting}   & \ding{51}  & 13.16 & 24.23 & 22.01 \\
    SCL~\cite{hou2022discovering} & \ding{51}  & 19.07 &30.39 &28.08 \\
    GEN-VLKT~\cite{liao2022gen}  &  \ding{51} & 21.36  & 32.91 & 30.56 \\
    OpenCat~\cite{zheng2023open}&  \ding{51} & 21.46  & 33.86 & 31.38 \\
    HOICLIP~\cite{ning2023hoiclip} & \ding{51} & \textbf{23.48}  & \textbf{34.47} & \textbf{32.26} \\
    \midrule
    \hline
    THID~\cite{wang2022learning}& \ding{55}  & 15.53 & {24.32} & {22.38} \\
    CMD-SE~\cite{lei2024exploring} & \ding{55} & {16.70} & 23.95 & 22.35 \\
     \textbf{\modelname} & \ding{55} & \textbf{23.27} & \textbf{28.34} & \textbf{27.22} \\
    \bottomrule
    \toprule[2pt]
  \end{tabular}
  \caption{Comparison of our proposed \modelname with state-of-the-art methods on the HICO-DET dataset.}
  \label{tab:hico-det}
\end{table}

\begin{table}[t]
  \centering
  \begin{tabular}{@{}lcccc@{}}
   \toprule[2pt]
    \toprule
    Method & Non-rare & Rare & Unseen & Full \\
    \midrule
    \midrule
    CHOID~\cite{wang2021discovering}  &   10.93  &   6.63  &   2.64  &   6.64\\
    QPIC~\cite{tamura2021qpic} &   16.95   &   10.84   &   6.21    &    11.12   \\
    GEN-VLKT~\cite{liao2022gen} &   20.91&   10.41 &   - &   10.87 \\
    MP-HOI~\cite{yang2024open}&   20.28 &   14.78 &   - &    12.61\\
    \midrule
    \hline
    THID~\cite{wang2022learning}  &   17.67   &   12.82   &   10.04   &    13.26   \\ 
    CMD-SE~\cite{lei2024exploring}           &  {21.46} & {14.64} & {10.70} & {15.26} \\
    \textbf{\modelname}    &  \textbf{23.67} & \textbf{16.55} & \textbf{12.46} & \textbf{17.20} \\
    \bottomrule
    \toprule[2pt]
    
  \end{tabular}
  \caption{Comparison of our proposed \modelname with state-of-the-art methods on the SWIG-HOI dataset.}
  \label{tab:swig-hoi}
\end{table}

\begin{table}[t]
  \centering
  \begin{tabular}{@{}ccccc@{}}
  \toprule[2pt]
    \toprule
      Method & Non-rare & Rare & Unseen & Full \\
    \midrule
    \midrule
        \textit{Base}   &  15.69  & 11.53  &  7.32  &  11.45 \\
    + \textit{GSA}  &  22.74 & 16.00   &  11.64 & 16.49  \\
    + \textit{HGC} &21.18  & 14.19  &  10.69 & 14.81 \\
    \bf{SGC-Net} &  \textbf{23.67} & \textbf{16.55} & \textbf{12.46} & \textbf{17.20} \\
     \bottomrule
     \toprule[2pt]
  \end{tabular}
  \caption{Ablation studies of the proposed method.}
  \vspace{-5mm}
  \label{tab:ablation-module}
\end{table}

\subsection{Comparisons with State-of-the-Art Methods}
\label{subsec:compare_sota}

\noindent \textbf{HICO-DET.}
To facilitate fair comparison, we first compare our model with state-of-the-art methods without pretrained detectors on the HICO-DET dataset. As shown in Table \ref{tab:hico-det}, SGC-Net outperforms CMD-SE~\cite{lei2024exploring} by 6.57\%, 4.39\%, and 4.87\% in mAP on the Unseen, Seen, and Full categories, respectively. Furthermore, even when compared to methods using pretrained detectors, SGC-Net achieves competitive performance. Although recent OV-HOI methods (\eg., GEN-VLKT~\cite{liao2022gen}, OpenCat~\cite{zheng2023open}, HOICLIP~\cite{ning2023hoiclip}) leverage CLIP text embeddings for interaction classification, they typically rely on a DETR architecture with pretrained weights. It is important to note that comparing our method with these approaches on HICO-DET dataset is not entirely fair, as the COCO dataset used for DETR pretraining shares the same object label space as HICO-DET.

\noindent \textbf{SWIG-HOI.}
Table~\ref{tab:swig-hoi} shows that SGC-Net outperforms all state-of-the-art methods on various setups. Specifically, SGC-Net outperforms CMD-SE~\cite{lei2024exploring} by 1.76\% and 1.91\% in mAP on the Rare and Unseen categories, respectively. Furthermore, SGC-Net outperforms the best OV-HOI method with a pretrained object detector by 4.59\% in mAP on the Full category. These results indicate that methods such as GEN-VLKT~\cite{liao2022gen} and MP-HOI~\cite{yang2024open}, which rely on pretrained detectors, perform suboptimally on the SWIG-HOI dataset. This is primarily because these methods struggle to scale effectively with vocabulary size, ultimately limiting their applicability in open-world scenarios. In contrast, SGC-Net overcomes this constraint by not relying on any detection pretraining, demonstrating superior capabilities in detecting and recognizing OV-HOI.

\begin{table}[t]
  \centering
   \resizebox{\columnwidth}{!}{%
   \begin{tabular}{@{}lcccc@{}}
   \toprule[2pt]
    \toprule
    Layer splitting manner & Non-rare & Rare & Unseen & Full \\
    \midrule
   
    \{4-6\}, \{7-9\}, \{10-12\}  & 23.03 & 15.72  & 11.35   & 16.35 \\
    \{6-8\}, \{9-10\}, \{11-12\}    & 23.61  & 16.52  &  11.64 & 17.01 \\ 
    \{6-8\}, \{9-11\}, \{12\}   &  \textbf{23.67} & \textbf{16.55} & \textbf{12.46} & \textbf{17.20}  \\
     \bottomrule
     \toprule[2pt]
   \end{tabular}%
  }
  \vspace{-3mm}
  \caption{Effectiveness of different layer splitting manners. The elements in \{\} means the layer numbers that are used.}
    \vspace{-1mm}
  \label{tab:ablation-last-layers}
\end{table}

\subsection{Ablation Study}
\label{subsec:ablation}

\begin{table}[t]
  \centering
  \resizebox{\columnwidth}{!}{%
    \begin{tabular}{lcccc}
      \toprule[2pt]
      \toprule
      Number of blocks & Non-rare & Rare & Unseen & Full \\
      \midrule
      \{12\} &21.18  & 14.19  &  10.09 & 14.81 \\
      \{9-11\}, \{12\}  & 21.91  & 14.48  & 10.01 & 14.78 \\
      \{6-8\}, \{9-11\}, \{12\} &   \textbf{23.67} & \textbf{16.55} & \textbf{12.46} & \textbf{17.20} \\
      \{3-5\}, \{6-8\}, \{9-11\}, \{12\}  & 22.61  & 15.25  & 12.01  & 16.53 \\

      \midrule
      \midrule
    Number of layers  &   &   &  &  \\
    \{8-9\}, \{10-11\}, \{12\}  & 22.45  &  15.17 & 11.55  & 15.98  \\
    \{6-8\}, \{9-11\}, \{12\}   &  \textbf{23.67} & \textbf{16.55} & \textbf{12.46} & \textbf{17.20}  \\

    \{4-7\}, \{8-11\}, \{12\}    &  23.50 & 16.21  &  12.13 &  16.75\\

      \bottomrule
      \toprule[2pt]
      
    \end{tabular}%
  }
    \vspace{-3mm}
  \caption{Evaluation on the number of blocks and layers.}
  \label{tab:ablation-groups-layers}
\end{table}

\begin{table}[t]
  \centering
   \begin{tabular}{@{}lcccc@{}}
   \toprule[2pt]
    \toprule
    Aggregation & Non-rare & Rare & Unseen & Full \\
    \midrule
    Self-attention & 21.37  &  13.43 &  7.88 &  13.90\\
    Concat  & 21.84 & 15.52 & 10.22 & 15.68 \\
    Sum           &22.09 &14.89 &9.59 &15.26  \\
    DGW           & 23.40  &  16.14 & 11.60  & 16.71 \\
    \midrule
    Sum*    &22.21 &15.16 &10.45 & 15.64\\
    DGW*  &  \textbf{23.67} & \textbf{16.55} & \textbf{12.46} & \textbf{17.20}  \\

     \bottomrule
     \toprule[2pt]
   \end{tabular}%
     \vspace{-2mm}
  \caption{Comparison of different aggregation methods. * denotes that the weights are trainable.}
  \vspace{-3mm}
  \label{tab:ablation-aggregation}
\end{table}

\begin{figure}[t]
\begin{center}
    \centering
    \includegraphics[scale=0.65]{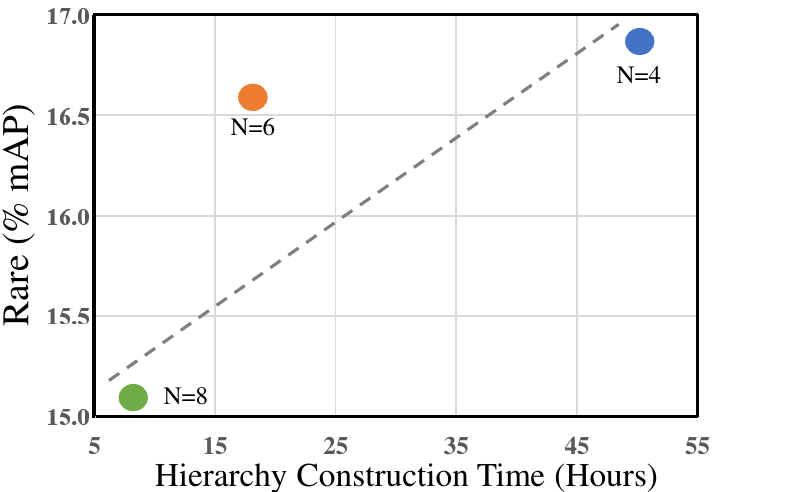}
    \vspace{-6mm}
    \captionof 
    {figure}{Evaluation on the value of grouping threshold $N$.}\label{tab:ablation-group}
    \vspace{-5mm}
\end{center}
\end{figure}

To prove the effectiveness of our proposed methods, we conduct five ablation studies on the SWIG-HOI dataset. Results of the ablation studies are summarized in Table~\ref{tab:ablation-module}-\ref{tab:ablation-aggregation}, and Figure~\ref{tab:ablation-group}, respectively.

\noindent \textbf{Effectiveness of the proposed modules.}
We first perform an ablation study to validate the effectiveness of the GSA and HGC modules. Results are summarized in Table~\ref{tab:ablation-module}. Our baseline model follows CMD-SE~\cite{lei2024exploring}, which utilizes CLIP and removes the need for pretrained object detectors in OV-HOI detection. First, we apply the GSA module on the baseline model to improve its transferability for novel classes, denoted as +\textit{GSA}. This modification achieves mAP improvements of 7.05\% for Non-Rare, 2.66\% for Rare, 4.32\% for Unseen, and 5.04\% for Full categories. Moreover, we integrate the HGC module with the baseline to enhance its ability to distinguish semantically similar interactions, denoted as +\textit{HGC}. The results indicate that the HGC module enables the baseline model to achieve 14.81\% in mAP for the Full categories. Finally, combining both modules results in the best performance, with an mAP of 12.46\% on the Unseen categories. These results demonstrate the significant potential of SGC-Net in enhancing interaction understanding within an open-vocabulary setting.

\begin{figure*}[t]
  \centering
  \hspace{5pt}
    \includegraphics[width=1.0\linewidth]{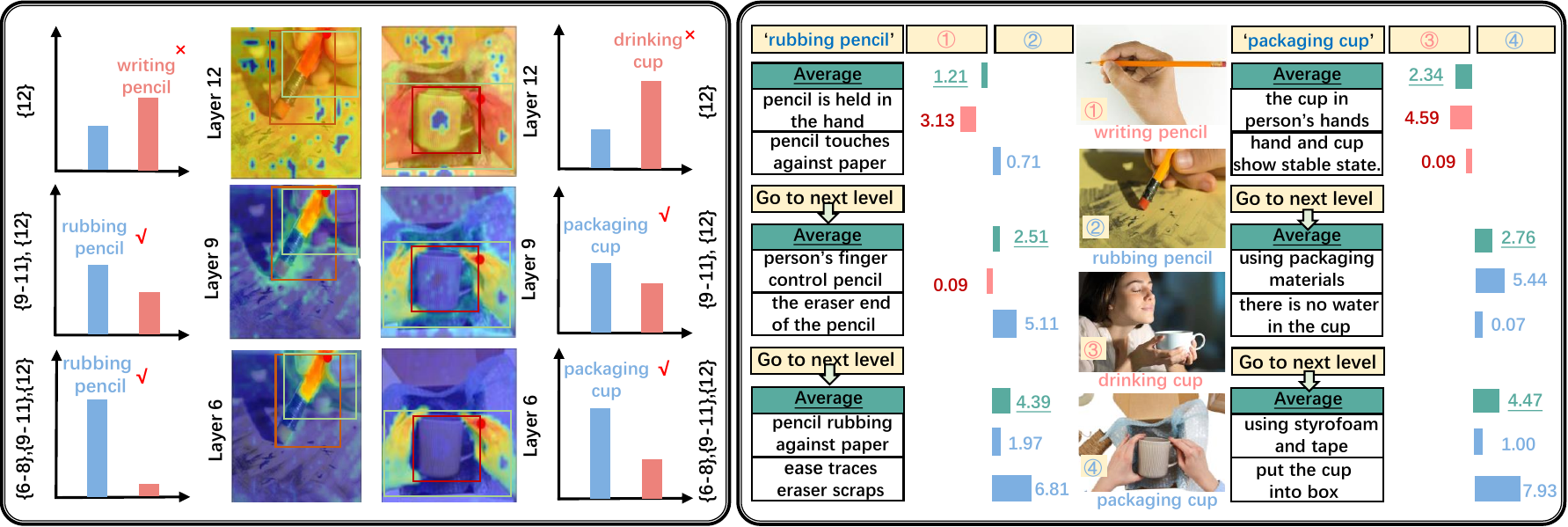}
    \vspace{-4.5mm}
   \caption{Qualitative results of SGC-Net. On the left-hand side, we visualize feature correspondence through cosine similarity calculations from both deep and shallow layers, and compare the predicted HOI categories. On the right-hand side, we show two inference examples and the absolute score gap between two images at each hierarchical level by querying the description with extracted HOI representations.}
    \vspace{-3.5mm}
   \label{fig:qual}
\end{figure*}

\noindent \textbf{Effectiveness of the proposed layer splitting manner.}
The block partitioning strategy plays a crucial role in the GSA module due to its potential to preserve visual-language associations. Experimental results in Table~\ref{tab:ablation-last-layers} indicate that separating the last layer into an independent block is more effective than other combinations of partitioning strategies. This is because the last layer of CLIP’s image encoder has the strongest association with text embeddings, and partitioning it into a separate block helps preserve this correlation while minimizing potential disruptions.

\noindent \textbf{Number of blocks and number of layers in each block.}
To show the importance of multi-granularity feature fusion across different layers, we compare the performance of SGC-Net with varying numbers of blocks and layers per block. As shown in Table~\ref{tab:ablation-groups-layers}, increasing the number of blocks from 1 to 3 gradually improves the performance, particularly for Unseen interactions. The best performance is achieved with three blocks, each containing three layers. Reducing the number of layers to 2 degrades performance, primarily due to neglecting mid-layer features. It should be noted that the beginning layers are excluded, as they primarily encode features with limited semantics.

\noindent \textbf{Comparison of different aggregation methods.}
Due to the substantial differences between shallow and deep features, directly integrating multi-granularity features may disrupt the alignment of vision-language embeddings in the pre-trained CLIP model. To address this issue, we introduce the DGW function, which mitigates these disruptions by accounting for the distances between layers. As shown in Table~\ref{tab:ablation-aggregation}, DGW consistently outperforms direct aggregation strategies such as Sum, Concat. Additionally, setting the aggregation weights as learnable further enhances the performance of DGW. These results demonstrate that our method, by assigning smaller weights to features from more distant layers during multi-granularity fusion, effectively improves model performance in open-vocabulary tasks.

\noindent \textbf{Grouping threshold $N$.}
A large grouping threshold, such as $N$ = 8, tends to produce rudimentary hierarchies and generalized descriptions. In contrast, smaller grouping thresholds increase the time required to construct the class hierarchy due to the more detailed recursive comparisons and groupings involved. As shown in Figure~\ref{tab:ablation-group}, the performance of SGC-Net improves as the grouping threshold increases. 
To balance computational efficiency and performance gains, we chose to set $N$ = 6.

\subsection{Qualitative Results}
To demonstrate the effectiveness of the proposed GSA module in leveraging CLIP’s multi-granularity features, we present cosine similarity maps of features and qualitative OV-HOI detection results. At the left hand side of Figure~\ref{fig:qual}, we show the patch similarity of Unseen classes not included in the training process. We observe that the intermediate layers of our method contain detailed information on local objects, including interaction patterns. Moreover, by leveraging these distinctive features, our SGC-Net improves HOI detection performance for both seen and unseen classes compared to using only last-layer features.

At the right hand side of Figure~\ref{fig:qual}, we show two groups of examples that demonstrate the inference and absolute score gap calculation via the proposed HGC module. Specifically, we input two Unseen class images (\,$\circlednumber{2}$ and $\circlednumber{4}$\,) and two Rare class images (\,$\circlednumber{1}$ and $\circlednumber{3}$\,) to evaluate their similarity with the hierarchical descriptions. The absolute score gap is computed as the absolute difference in similarity scores between paired images: \ie, $\circlednumber{1}$ vs. $\circlednumber{2}$ and $\circlednumber{3}$ vs. $\circlednumber{4}$.  Due to the weak comparative information in the early descriptions, similar images tend to yield similar scores at the beginning. As we progress towards a more comparative description, the disparity between the scores of the two images gradually increases. Besides, this also highlights the interpretability of our method, as it provides the user insights regarding the attributes that elicit stronger responses to the input. Additionally, it indicates the specific layer of description at which the scores of different images diverge, widening the gap.

\section{Conclusion}
\label{sec:conclusion}

In this paper, we propose the SGC-Net which enhances CLIP's transferability and discriminability to handle two critical issues in OV-HOI: feature granularity deficiency and semantic similarity confusion. Specifically, the GSA module enhances CLIP's image encoder by improving its ability to capture fine-grained HOI features while maintaining alignment between multi-granularity visual features and text embeddings. The HGC module strengthens CLIP's text encoder in distinguishing semantically related HOI categories by incorporating a grouping and comparison strategy into the LMM. Through extensive comparative experiments and ablation studies, we validate the effectiveness of SGC-Net on two datasets.

{
    \small
    \bibliographystyle{ieeenat_fullname}
    \bibliography{main}
}


\end{document}